\documentclass[10pt,onecolumn]{IEEEtran}
\usepackage{amsfonts}
\usepackage{indentfirst}
\usepackage{amsmath}
\usepackage{amsthm}
\usepackage{amssymb}
\usepackage{color}
\usepackage{graphicx}
\usepackage[para]{threeparttable}
\usepackage{booktabs}
\usepackage{caption}
\usepackage{makecell}
\usepackage{diagbox}
\usepackage{multirow}
\usepackage{epstopdf}
\usepackage{floatrow}
\usepackage{booktabs}
\usepackage{subfigure}

\usepackage{algorithm}
\usepackage{algpseudocode}
\usepackage{bm}

\normalsize

\allowdisplaybreaks[4]

\begin{document}
\title{Convolutional Neural Network with Convolutional Block Attention Module for Finger Vein Recognition}

\author{Zhongxia Zhang, Mingwen Wang
\thanks{Z. Zhang and M. Wang are with  School of Mathematics, Southwest Jiaotong University, Chengdu, 610031, China. E-mail: 1065947699@qq.com, wangmw@swjtu.edu.cn.}
\thanks{This work was partially supported by the Fundamental Research Funds for the Central Universities under Grant 2682021ZTPY100, in part by the Science and Technology Support Project of Sichuan Province under Grant 2020YFG0045 and 2020YFG0238.}
\thanks{Corresponding author: M. W. Wang.}
}
\markboth{IEEE Transactions on Instrumentation and Measurement}{Submitted paper}
\maketitle
\begin{abstract}
Convolutional neural networks have become a popular research in the field of finger vein recognition because of their powerful image feature representation. However, most researchers focus on improving the performance of the network by increasing the CNN depth and width, which often requires high computational effort. Moreover, we can notice that not only the importance of pixels in different channels is different, but also the importance of pixels in different positions of the same channel is different. To reduce the computational effort and to take into account the different importance of pixels, we propose a lightweight convolutional neural network with a convolutional block attention module (CBAM) for finger vein recognition, which can achieve a more accurate capture of visual structures through an attention mechanism. First, image sequences are fed into a lightweight convolutional neural network we designed to improve visual features. Afterwards, it learns to assign feature weights in an adaptive manner with the help of a convolutional block attention module. The experiments are carried out on two publicly available databases and the results demonstrate that the proposed method achieves a stable, highly accurate, and robust performance in multimodal finger recognition.
\end{abstract}


\IEEEpeerreviewmaketitle
\section{Introduction}
With the development of society, people have higher and higher requirements for identity information security. Traditional identification technology has been difficult to meet people's needs, so it is necessary to develop biometric technology with higher security. Finger veins are hidden under the skin of the fingers \cite{A-Kumar-2012},\cite{H-Y-2020}. Compared with other biometric features, its structure is complex and not available under visible light. It has high stability, concealment and anti-counterfeiting properties and has a broad application prospect.

Generally speaking, the finger vein recognition process includes the following four steps: image acquisition, pre-processing, feature extraction and matching \cite{J-D-Wu-2009}. Among them, feature extraction plays a crucial role. According to the different feature extraction methods, the existing finger vein feature extraction methods can be roughly divided into two groups: vein pattern-based methods and local binary-based methods. These methods perform well in most cases.

Relative to the above-mentioned custom features, deep features learned from convolutional deep neural networks have been shown to have better generalization and representation \cite{J-D-Wu-2011},\cite{B-Hou-2020}. Recent studies have shown that deep convolutional neural networks (CNNs) have outstanding performance in the field of image understanding and recognition \cite{C-Xie-2019},\cite{Z-M-2019}, which motivated us to employ deep CNNs for feature learning and selection of finger vein images \cite{J-Y-2020}. For example, Hong \emph{et al.} \cite{H-G-Hong-2017} applied a pre-trained CNN model of VGG-Net-16 for finger vein validation. Yin \emph{et al.} \cite{Y-Yin-2011} fine-tuned VGG-Net-16 and achieved good recognition accuracy on two publicly available databases of finger veins. Besides, there are many studies \cite{F-J-Xu-2017},\cite{S-Y-Li-2021},\cite{{Y-Fang-2018}} devoted to improve the results by increasing the depth or width of neural networks, although all of them achieved good results, but the network is too deep or too wide, which is a great test for the computer's computational power.

It is known that an important feature of the human visual system is that it does not try to process the whole scene immediately, but selectively focuses on salient parts in order to better capture the visual structure. Attention can be directed to focus, and expressivity can be improved by using attentional mechanisms, i.e., focusing on important features and suppressing unnecessary ones. The convolutional block attention module (CBAM) \cite{S-Woo-2018} is a simple and effective attention module for feedforward convolutional neural networks that can attend to important information in channels and spaces separately, which not only saves parameters and computational power, but also ensures its integration into existing network architectures as a plug-and-play module.

Motivated by the success of CNN and CBAM, we propose a novel and effective feature representation of lightweight CNN based on CBAM blocks for the finger recognition to save computational power. By embedding CBAM in the designed lightweight CNN architecture, the accuracy is improved while ensuring small computational power. Specifically, the image is fed into a lightweight base CNN architecture, and the initial features of the finger vein are extracted using the powerful feature representation capability of CNN. Then, CBAM blocks are embedded in the original network to infer the attention mapping according to two independent dimensions - channel and spatial order, and the attention mapping is multiplied into the input feature mapping with adaptive feature refinement. Finally, the output features are classified using the softmax function, and the method proposed in this paper has significant improvements in two different databases. The main contributions are as follows:
\begin{itemize}
  \item[(1)] To our knowledge, this paper is the first to successfully apply lightweight CBAM blocks into CNN for finger vein recognition.

  \item[(2)] We combine a convolutional block attention module (CBAM) with a lightweight CNN to simulate visual attention mechanisms and enhance the flow of information in channels and spaces.

  \item[(3)] Our network structure is simple and guarantees the smallest possible computation without sacrificing network performance. Experimental results show that the scheme in this paper has competitive potential in finger vein recognition systems.
\end{itemize}

The rest of the paper is organized as follows. Section II is to describe the related work. Section III presents the theoretical background of CNN and CBAM. Proposed approach is discussed in Section IV. In Section V, we present experimental results. Finally, Section VI concludes the paper.
\section{Related work}
Undoubtedly, a proper feature extraction method in finger vein recognition systems can be of great benefit in improving the performance \cite{L-Yang-2018},\cite{H-C-2010}. How to effectively extract discriminative features remains a major challenge for finger recognition. The existing finger feature extraction methods can be broadly classified into two categories \cite{K-X-2021}: traditional recognition methods and deep learning methods.

Traditional recognition methods generally include subspace learning-based methods, vein pattern-based, detail point matching-based, and local feature-based methods.
\begin{itemize}
  \item[(1)] Subspace learning approaches are based on machine learning methods to reduce the dimensionality of global features and filter noise at the same time, such as principal component analysis (PCA) \cite{J-Wang-2013}, linear discriminant analysis (LDA) \cite{J-Wu-2011} and sparse representation (SR) \cite{Y-Xin-2012}, all of which transform texture images into different subspaces and generate feature vectors from individual coefficients of the subspace to accomplish texture recognition. Wang \emph{et al.} \cite{J-Wang-2013} combined the traditional 2D-PCA and 2D-FLD (Fisher linear discriminant) techniques, Wu and Liu \cite{J-Wu-2011} implemented finger vein classification using PCA and LDA, and Xin \emph{et al.} \cite{Y-Xin-2012} successfully applied SR to the finger vein recognition task. These PCA, LDA and SR-based methods can reduce the preprocessing steps and have a smaller space occupation of the feature vector. However, they extract features from a global perspective and do not provide sufficient description of local feature information.
  \item[(2)] Vein-based methods segment the vein pattern from the finger vein image and match it by geometry or topology. Typical methods include the mean curvature method \cite{W-Song-2011}, the maximum curvature (MC) point method \cite{N-Miura-2007}, the repetitive line tracing method (RLT) \cite{N-Miura-2004}, and morphological operations combined with the mean Gabor method \cite{A-Kumar-2012},\cite{J-Yang-2012},\cite{W-Y-2012}, etc. The RLT \cite{N-Miura-2004} method extracts the vein network by calculating the difference between the central pixel value and the pixel value in the corresponding range, which has high time complexity and does not take into account the symmetry and continuity of the vein pattern. Miura \emph{et al.} \cite{N-Miura-2007} modified the RLT method by calculating the maximum curvature information during vein tracing. Subsequently, both literature \cite{W-Song-2011} and \cite{H-Qin-2017} also used the curvature method to extract finger vein features. Kumar \emph{et al.} \cite{A-Kumar-2012} used Gabor filter to extract vein patterns. Although Gabor filter is powerful in image texture analysis, it causes information loss for low quality vein images. Recently, Yang \emph{et al.} \cite{L-Yang-2019} proposed a finger vein code indexing method and combined it with finger vein pattern matching method into an integrated framework. And the final experimental results show that the integrated framework does improve the recognition accuracy. However, the method, like other vein-based methods, requires segmentation of finger vein images, which is greatly affected by the quality of finger images.
  \item[(3)] Detail point matching-based methods extract stable structures such as intersections and endpoints of vessels as feature points to calculate the similarity of two matched images, including the improved Hausdorff distance matching method \cite{C-Yu-2009}, the singular value decomposition-based detail matching method \cite{F-Liu-2014}. Besides, the scale-invariant feature transform (SIFT)-based method \cite{H-Kim-2012},\cite{S-Pang-2012} can automatically extract feature points from finger vein images for matching, and is usually regarded as a method based on minutiae points. Recently, Meng \emph{et al.} \cite{X-Meng-2021} combined detail point matching with the traditional region of interest (ROI)-based method to select details of reasonable neighborhoods for matching, which avoids mismatch to some extent and is more stable in detail matching. Use of above features for vein matching usually shows poor performance for low-quality images due to the presence of spurious minutiae features. In addition, employing minutiae features are  sensitive to changes in finger pose \cite{J-Peng-2014}.
  \item[(4)] Local feature-based methods, such as local binary patterns (LBP) \cite{L-Yang-2014},\cite{C-Liu-2016}, local line binary patterns (LLBP) \cite{B-A-2011}, local line directional patterns (LLDP) \cite{Y-T-2016}, and local graph structures (LGS) \cite{M-F-2014},\cite{S-Dong-2015}, have been widely used in finger biometrics. For example, Dong \emph{et al.} \cite{S-Dong-2015} proposed a multi-directional weighted symmetric local graph structure (MoW-SLGS) operator for finger vein recognition. Unfortunately, the operator assigns different weights to symmetric pixels, which leads to an unbalanced feature representation between the left and right sides. Liu \emph{et al.} \cite{Y-Liu-2013} stated a multi-directional local line binary pattern (PLLBP) method that makes full use of the discriminative power of the LLBP histogram in different directions. And the method can be used to extract vein line patterns in any direction. Recently, Al-Nima \emph{et al.} \cite{A-Nima-2017} proposed the multi-scale Sobel angular local binary pattern (MSALBP) for the feature extraction algorithm of finger texture images. This method combines the Sobel operator with a multi-scale local binary pattern, which is computed statistically for each image block to form a texture vector as an input to an artificial neural network (ANN). In general, although these local feature-based methods can better capture the local information of the image, the methods only consider the relationship between the target pixel and its surrounding pixels, ignoring the hidden relationship between surrounding neighboring pixels. In other words, the uniqueness of finger images will not be effectively expressed by relying only on hand-crafted features.
\end{itemize}

Although the traditional recognition method has achieved good results \cite{S-Li-2021}, but its recognition process is more complicated, and the extracted features are the shallow features of the image, which are easily affected by the image quality. Deep learning methods are not easily affected by image quality, have powerful image processing without any prior knowledge, and perform well in noisy image processing and adaptive learning of feature representations. In the field of finger vein recognition, deep learning method has been successfully applied in recent years \cite{F-J-2017}. For example, Huang \emph{et al.} \cite{H-Huang-2017} designed a new finger vein verification method with deep convolutional neural network, which works well for finger vein pattern matching. In \cite{R-Das-2019}, Das \emph{et al.} developed a CNN-based finger-vein identification system, which performed an effective identification, but did not consider factors such as training time and model size. Beisides, Ahmad Radzi \emph{et al.} \cite{S-A-2016} used a four-layer CNN to recognize finger veins. The recognition accuracy reached $100\%$, but the results were obtained in their own database, which is not universal. And Li \emph{et al.} \cite{R-Li-2020} used improved GCNs to recognize finger veins and obtained better recognition accuracy, but image preprocessing was needed to construct a weighted map of finger veins. He \emph{et al.} \cite{X-He-2019} enhanced the feature extraction ability of the network by adding convolution layer, which improved the recognition accuracy with less training samples, and also increased the complexity of the network.
\section{Methodology}
\begin{figure*}
  \centering
  \includegraphics[width=14cm]{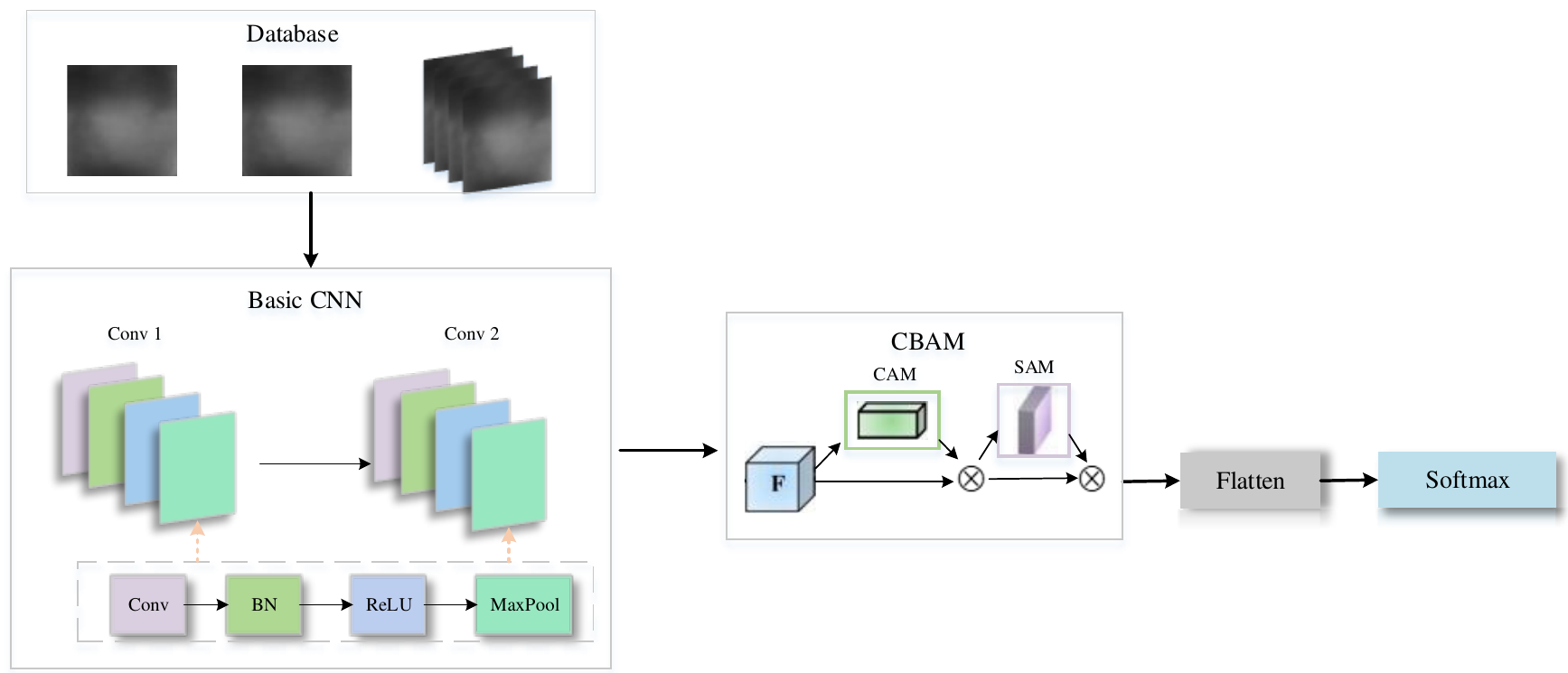}
  \caption{General framework of the network for the proposed approach}\label{fig-5}
\end{figure*}

CNN has been successfully applied in the field of computer vision, demonstrating its powerful ability to represent features. It uses multiple filters that share different parameters to extract image features. Recently, the attention mechanism has received increasing attention for their focus on important details and their ability to better capture visual structure. In our study, a CNN with a convolutional attention module (CBAM) is proposed. It uses a basic CNN to extract the overall features of all input data. Then, the CBAM processes the feature map and adaptively assigns weights to the channel dimensions and spatial dimensions. Both the basic CNN architecture and CBAM are indispensable because the basic network architecture focuses on the global information, while CBAM highlights its features. These two parts complement each other and complete the neural network. Fig. \ref{fig-5} shows the complete framework of the proposed network, and we explain the components of the framework (i.e., the basic CNN, CBAM and output layer) in detail below.

\subsection{The Basic CNN}
 In our work, considering the problem of overfitting when training with small data in a too deep network, we designed a lightweight CNN with the network parameters set as shown in Table \ref{ta-net-str}. The lightweight CNN consists of a total of two convolutional modules, each with a convolutional layer, a normalization layer, an activation function layer, and a pooling layer.
 \begin{table}[!h]
\centering
\caption{Details of the proposed network framework} \label{ta-net-str}
\begin{tabular}{cccc}
  \toprule
  Layer Type & \makecell{Number of\\ filters} & Size & Output size\\
  \midrule
  Input      & $-$ & $-$ & $1\times81\times333$ \\
  Conv 1     & $16$ & $5\times5$ & $16\times81\times333$ \\
  BN     & $1$ & $-$ & $16\times81\times333$ \\
  ReLU     & $1$ & $-$ & $16\times81\times333$ \\
  Pool 1     & $1$ & $3\times3$ & $16\times27\times111$ \\
  Conv 2     & $32$ & $5\times5$ & $32\times27\times111$ \\
  BN     & $1$ & $-$ & $32\times27\times111$ \\
  ReLU     & $1$ & $-$ & $32\times27\times111$ \\
  Pool 2     & $1$ & $3\times3$ & $32\times9\times37$ \\
  \bottomrule
\end{tabular}
\end{table}

For convolutional layer, features are extracted by performing a two-dimensional convolution of the input graph and the convolutional kernel, which can be specifically expressed by Eq. (\ref{eq-1}) as:
\begin{align}\label{eq-1}
  y_1 =\sigma(\mathbf{b}+\sum_{l=0}^{k-1}\sum_{m=0}^{k-1}\mathbf{w}_{l,m}\mathbf{a}_{i+l,j+m})
\end{align}
where $\sigma$ represents the activation function, such as sigmoid function, etc.; $\mathbf{b}$ is the offset value; $\mathbf{w}$ is the $k\times k$ size shared weight matrix. We use matrix $\mathbf{a}$ to represent the input layer neurons,  and $\mathbf{a}_{x,y}$ to denote the neuron in the $(x+1)$-th row and $(y+1)$-st column (note that the subscripts here are counted from 0 by default, $\mathbf{a}_{0,0}$ represents the neuron in the first row and first column), so we get Eq. (\ref{eq-1}) by adding the offset value after the matrix  $\mathbf{w}$ linear mapping.

The batch normalization (BN) layer normalizes the feature maps generated by the convolutional layers and sends these feature maps to activation function to speed up the training process. As for the activation function, rectified linear units (ReLU) were chosen in Eq. (\ref{eq-1}). ReLU is introduced as the activation function of the hidden layer instead of the traditional sigmoid unit because ReLU is better at capturing patterns in natural images and improves the ability of the neural network to solve the image denoising problem.

The pooling layer compresses the information in the original feature layer so that the input representation can be more compact. In general, there are two operations: maximum pooling and average pooling. Max-pooling simply reduces the dimensionality of the data by taking only the largest data, while average-pooling works in a similar way by taking the average of the inputs instead of the maximum. Based on the conceptual differences between these two methods, max-pooling is sensitive to the texture information of the image, while the average pooling method retains more background information of the image. Therefore, max-pooling is more beneficial for extracting the feature information of the image. In this paper, max-pooling is used after calculating the ReLU output.
\subsection{Convolutional Block Attention Module (CBAM)}
\begin{figure}[!h]
  \centering
  \includegraphics[width=8cm]{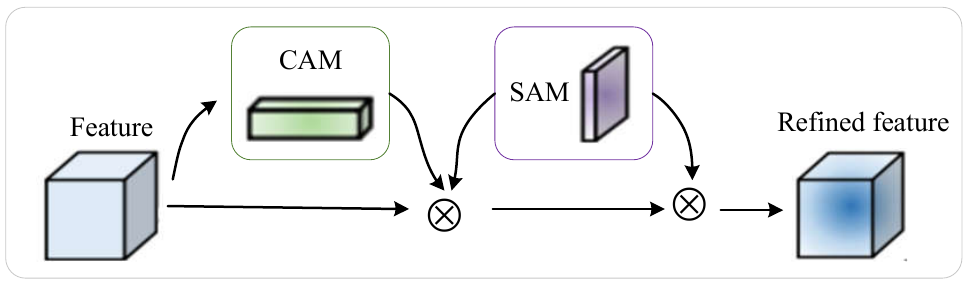}
  \caption{Convolutional block attention module}\label{fig-2}
\end{figure}

The above basic CNN network transforms the image data $\mathbf{X}$ into a feature map $\mathbf{F}$ as the input to CBAM. The CBAM contains two independent sub-modules, channel attention module (CAM) and spatial attention module (SAM), which can save parameters and computational power by performing attention mechanism on channel and space respectively, as shown in Fig. \ref{fig-2}.

CBAM inferred the attention maps one at a time along two independent dimensions (channel and space) and then multiplied the attention maps by the input feature maps for adaptive feature refinement. The process of  the input feature map $\mathbf{F}$ being processed in CBAM can be summarized as:
\begin{align}\label{eq-2}\begin{split}
  \mathbf{F}'=&\mathbf{M_{c}}(\mathbf{F})\otimes\mathbf{F} \\
  \mathbf{F}''=&\mathbf{M_{s}}(\mathbf{F}')\otimes\mathbf{F}'
\end{split}\end{align}
where $\otimes$ denotes element multiplication, $\mathbf{F}'$ denotes the result of multiplying the feature map with the channel attention map, and $\mathbf{F}''$ is the final refined output.

Below, we will detail how the two separate dimensions, CAM and SAM, work.
\subsubsection{CAM}
\begin{figure}[!h]
  \centering
  \includegraphics[width=8cm]{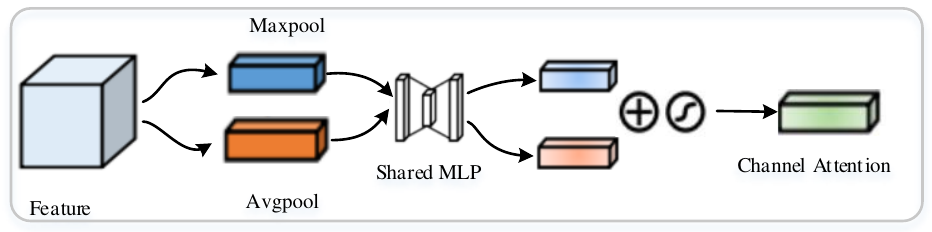}
  \caption{Channel attention module}\label{fig-3}
\end{figure}

To accomplish feature extraction and reduce data loss, the channel attention module compresses the feature maps on the spatial dimension using the global average pooling layer and the global maximum pooling layer. Fig. \ref{fig-3} shows the channel attention module. The global average pooling layer obtains the overall information, while the global maximum pooling layer obtains the feature variance information. The combination of these two layers is better than any layer.

The compressed $\mathbf{F_{avg}^{C}}$ and $\mathbf{F_{max}^{C}}$ descriptors are then sent to a shared network to generate our channel attention map $\mathbf{M_{c}}\in \mathbb{R}^{C\times 1\times 1}$. The shared network consists of a multi-layer perceptron (MLP) with a hidden layer. To reduce the parameter overhead, the hidden activation size is set to $\mathbb{R}^{C/r\times 1\times 1}$, where $r$ is the reduction rate. After the shared network is applied to each descriptor, we use element summation to merge the output feature vectors. In short, the channel attention is calculated as Eq. (\ref{eq-3}):
\begin{align}\label{eq-3}\begin{split}
  \mathbf{M_{c}}(\mathbf{F})&=\sigma(MLP(AvgPool(\mathbf{F}))+MLP(MaxPool(\mathbf{F}))) \\
   &=\sigma(\mathbf{W}_{1}(\mathbf{W}_{0}(\mathbf{F_{avg}^{C}}))+\mathbf{W}_{1}(\mathbf{W}_{0}(\mathbf{F_{max}^{C}})))
\end{split}\end{align}
where $\sigma$ denotes the sigmoid function, $\mathbf{W}_{0}\in \mathbb{R}^{C/r\times C}$, and $\mathbf{W}_{1}\in \mathbb{R}^{C\times C/r}$. Note that the MLP weights $\mathbf{W}_{0}$ and $\mathbf{W}_{1}$ are shared between the two inputs, and the ReLU activation function is followed by $\mathbf{W}_{0}$.
\subsubsection{SAM}
\begin{figure}[!h]
  \centering
  \includegraphics[width=8cm]{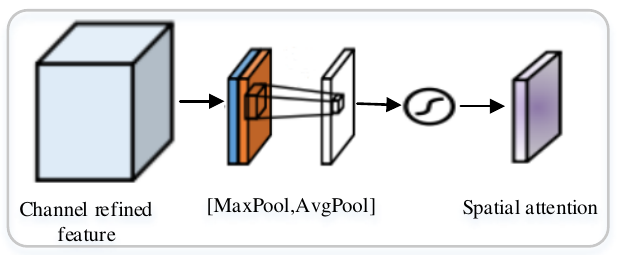}
  \caption{Spatial attention module}\label{fig-4}
\end{figure}

The feature map $\mathbf{F}'$ output from CAM is used as the input feature map of this module. First, we perform global max pooling and global average pooling based on channels to obtain two $H\times W\times1$ feature maps, and then we do concat operation (channel splicing) on these two feature maps based on channels. Next, after a $7\times7$ convolution ($7\times7$  is better than $3\times3$ ) operation, it is reduced to one channel, i.e., $H\times W\times1$. After that, the spatial attention feature is generated by sigmoid function. Finally, this feature is multiplied with the input feature of the module to get the final generated feature. Specifically, as shown in Fig. \ref{fig-4}, the calculation process is as follows Eq. (\ref{eq-4}):
\begin{align}\label{eq-4}\begin{split}
  \mathbf{M_{s}}(\mathbf{F})&=\sigma(f^{7\times 7}([AvgPool(\mathbf{F});MaxPool(\mathbf{F})])) \\
   &=\sigma(f^{7\times 7}[\mathbf{F_{avg}^{s}};\mathbf{F_{max}^{s}}])
\end{split}\end{align}
where $\sigma$ is a sigmoid function and $f^{7\times 7}$ denotes a convolution operation with a filter size of $7\times7$.

CAM utilizes the inter-channel relationships of features to generate a channel attention map. As each channel of a feature map is considered as a feature detector, channel attention focuses on `what' is meaningful given an input. SAM uses spatial relationships between features to generate spatial attention maps. Unlike CAM, spatial attention focuses on the `where' is an informative part, which is complementary to CAM.
\subsection{Output layer}
The feature map $\mathbf{F}$ is refined into $\mathbf{F}''$ after going through the CAM and SAM modules. $\mathbf{F}''$ is the final feature obtained from our proposed network framework. Additional dense layers are needed in our network to recreate the data structure used for classification. Softmax is used as the final activation because it is a sigmoid equivalent, however with traditionally better results and a normalized output essential for classification problems with multiple classes.

The definition of the Softmax function (with the $i$-th node output as an example) is given below:
\begin{align}\label{eq-12}
  Softmax(z_{i})=\frac{e^{z_{i}}}{\sum_{c=1}^{C}e^{z_{c}}}
\end{align}
where $z_{i}$ is the output value of the $i$-th node and $C$ is the number of output nodes, i.e., the number of categories of the classification. The Softmax function allows converting the output values of multiple categories into a probability distribution ranging from [0, 1] and summing to 1. In classification problems, we want the model to assign probabilities close to 1 to the correct category and 0 to the others, which is difficult to achieve if we use linear normalization methods. Softmax has a `two-step' strategy of discretization followed by normalization, so it has a significant advantage in classification problems.
\section{Experimental Results and Analysis}
\subsection{Database}
In this section, we will discuss our experiments to confirm the effectiveness of our proposed method and show the observed results. In our experiment, we used vein images from two public finger vein databases: the Hong Kong Polytechnic University database (HKPU) \cite{A-Kumar-2012} and the University Sains Malaysia database (USM) \cite{M-S-2014}. The characteristics of the databases used in our experiment are shown in Table \ref{ta-detail-database}.The details of these databases are given below:
\begin{table}[!h]
\centering
\caption{The details of databases} \label{ta-detail-database}
\begin{tabular}{ccccc}
  \toprule
  Database & \makecell{Finger \\number} & \makecell{Image number\\per finger} & \makecell{Size of raw\\image (pixels)} & ROI image \\
  \midrule
  HKPU & $312$ & $6$ & $513\times256$ & \makecell{Extract using\\ the method of  \cite{A-Kumar-2012}}\\
  USM & $492$ & $12$ & $640\times480$ & \makecell{Database includes\\ ROI images}\\
  \bottomrule
\end{tabular}
\end{table}

(1) HKPU Database: The images of the database were collected from 156 volunteers. Every volunteer only collected the index finger and middle finger of his left hand, totaling 312 fingers. Each finger collected 6 images. Each of first 210 fingers has 12 images, captured in two separate sessions, and others each has 6 images, captured in one session. All of the images are 8-bit gray level BMP files with a resolution of $513\times256$ pixels. In our experiments, we use 6 samples each fingers and a total of 312 finger types were obtained. We used the method in the literature \cite{A-Kumar-2012} to perform ROI operations on the original images and obtained an image size of $333\times 81$.

(2) USM Database: The database of Universiti Sains Malaysia (USM) consists of 492 fingers, and every finger provided 12 images. The spatial resolution of the captured finger images were $640\times 480$. This database also provides the extracted ROI images with a size of $300\times100$ for finger vein recognition using their proposed algorithm described in \cite{M-S-2014}.
\subsection{Parameter Settings and Experiments}
In our method, we adopt accuracy as a measure of model performance, the formula is as follows:
\begin{align}\label{eq-5}
 Accuracy=\frac{\text{Total number of correct identifications}}{\text{Total number of samples}}
\end{align}

We implement the proposed method using Pytorch framework and conduct the experiments on a common desktop PC with i7 3.60 GHz CPU. In the proposed model, Adam's method is chosen as the optimizer considering the performance and training time because of its good performance in improving traditional gradient descent and facilitating hyperparameter dynamic tuning. The learning rate of the model was set to $0.0001$ and the batch size was 36. In the next experiments, the number of epochs was defined as $20$. For training purposes, $70\%$ of the finger vein images were randomly selected as the original training samples and the remaining $30\%$ were considered as the test samples. As a rough estimate because the proposed networks are lightweight and easily trained, it takes less than half an hour to train.
\subsection{Experiments}
\begin{figure}[!t]
\centering
\subfigure[]{
\begin{minipage}[t]{0.4\linewidth}
\centering
\includegraphics[width=7cm]{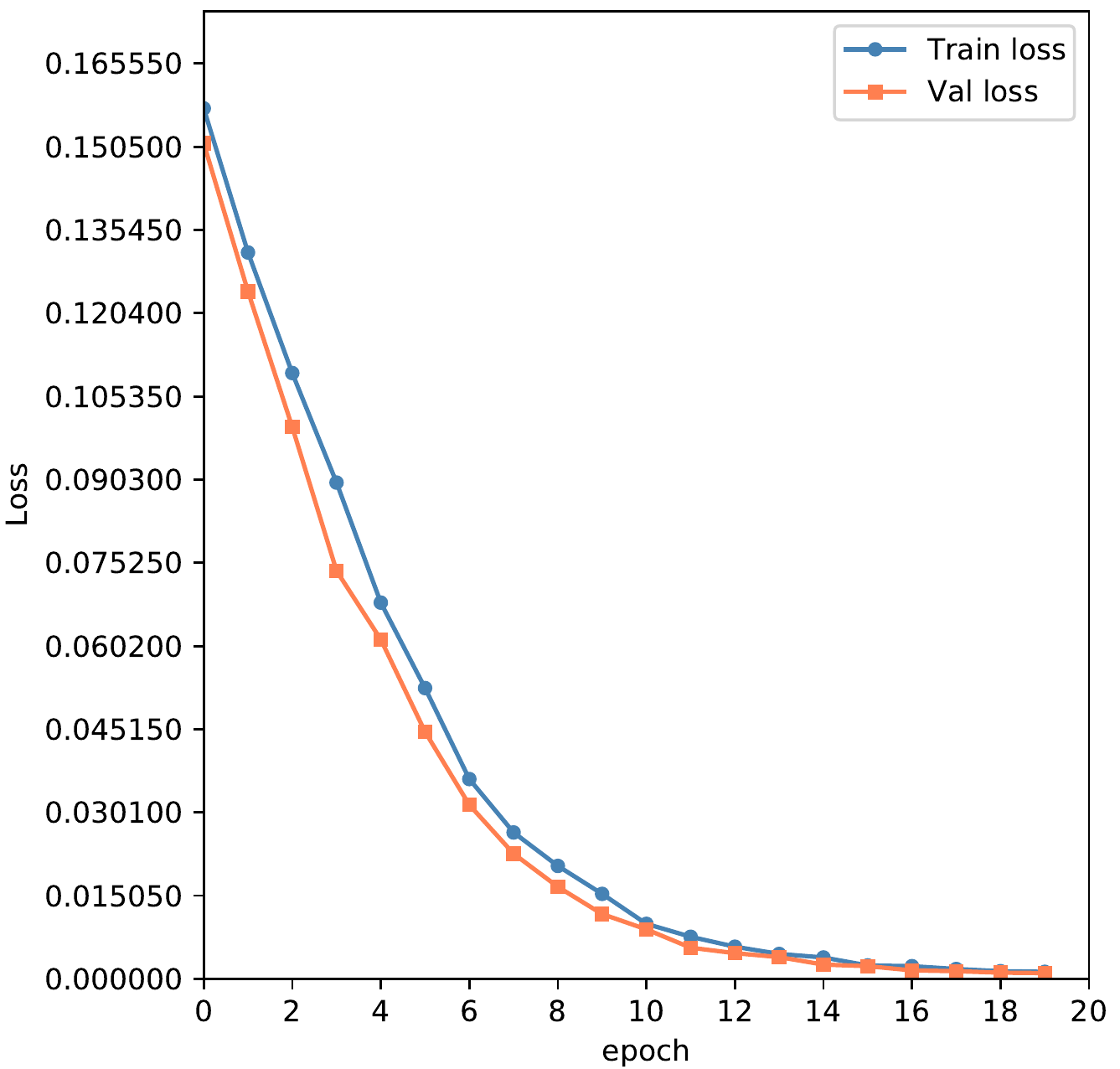} \label{fig-ha}
\end{minipage}
}
\subfigure[]{
\begin{minipage}[t]{0.4\linewidth}
\centering
\includegraphics[width=7cm]{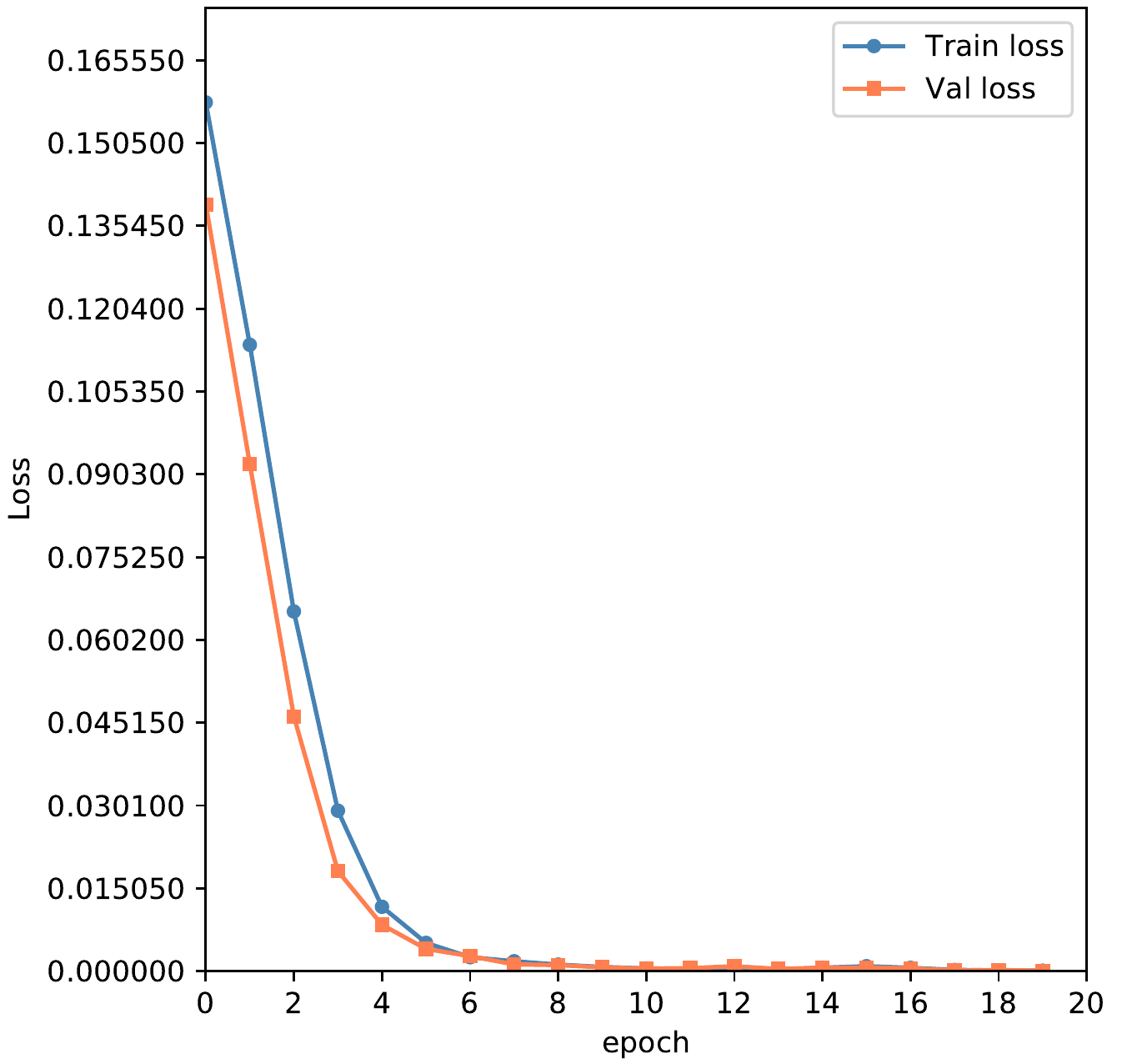} \label{fig-ua}
\end{minipage}
}
\caption{Loss curves on two public databases: (a) HKPU, (b) USM.} \label{fig-loss}
\end{figure}

\begin{figure}[!t]
\centering
\subfigure[]{
\begin{minipage}[t]{0.4\linewidth}
\centering
\includegraphics[width=7cm]{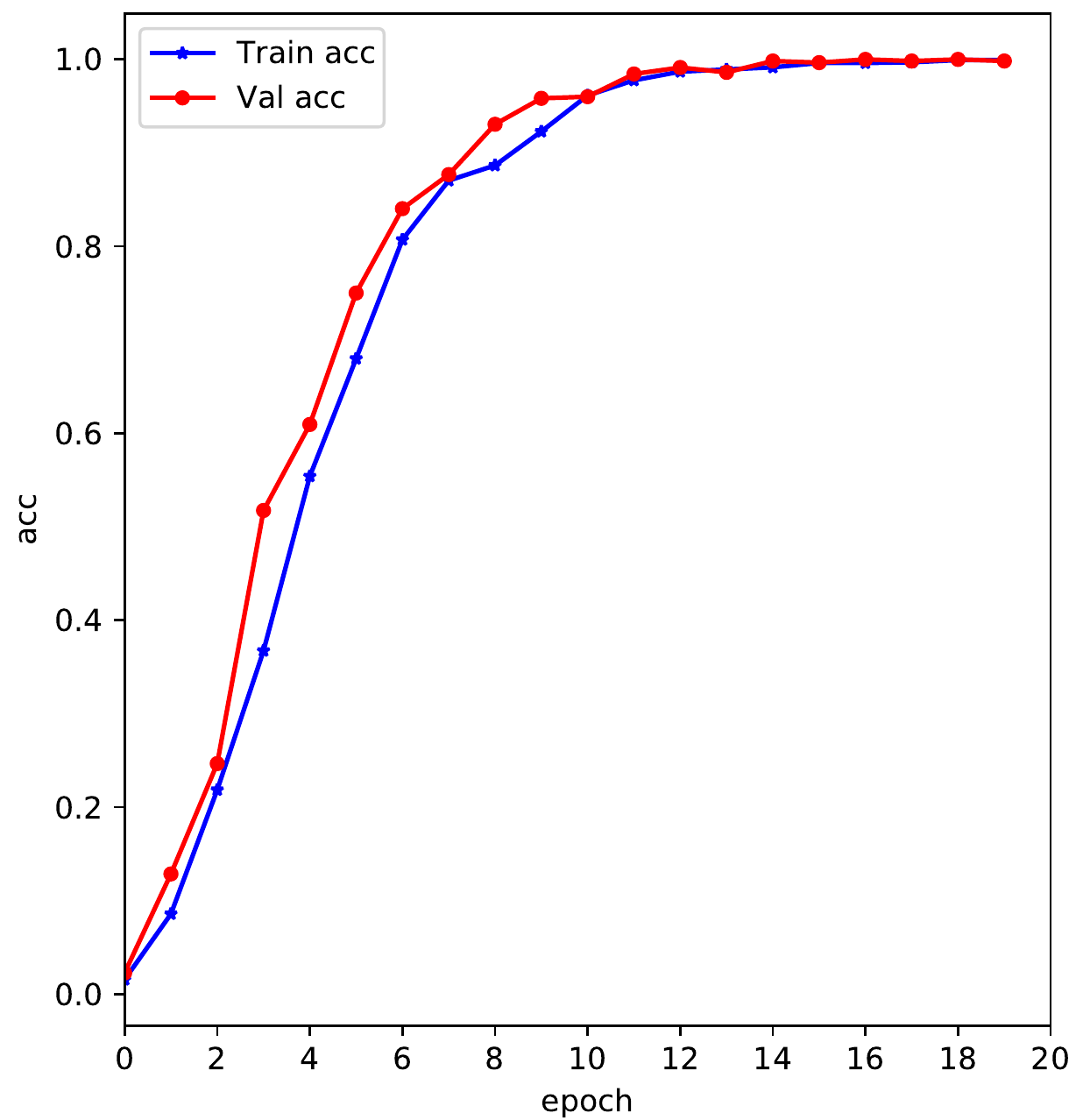} \label{fig-hb}
\end{minipage}
}
\subfigure[]{
\begin{minipage}[t]{0.4\linewidth}
\centering
\includegraphics[width=7cm]{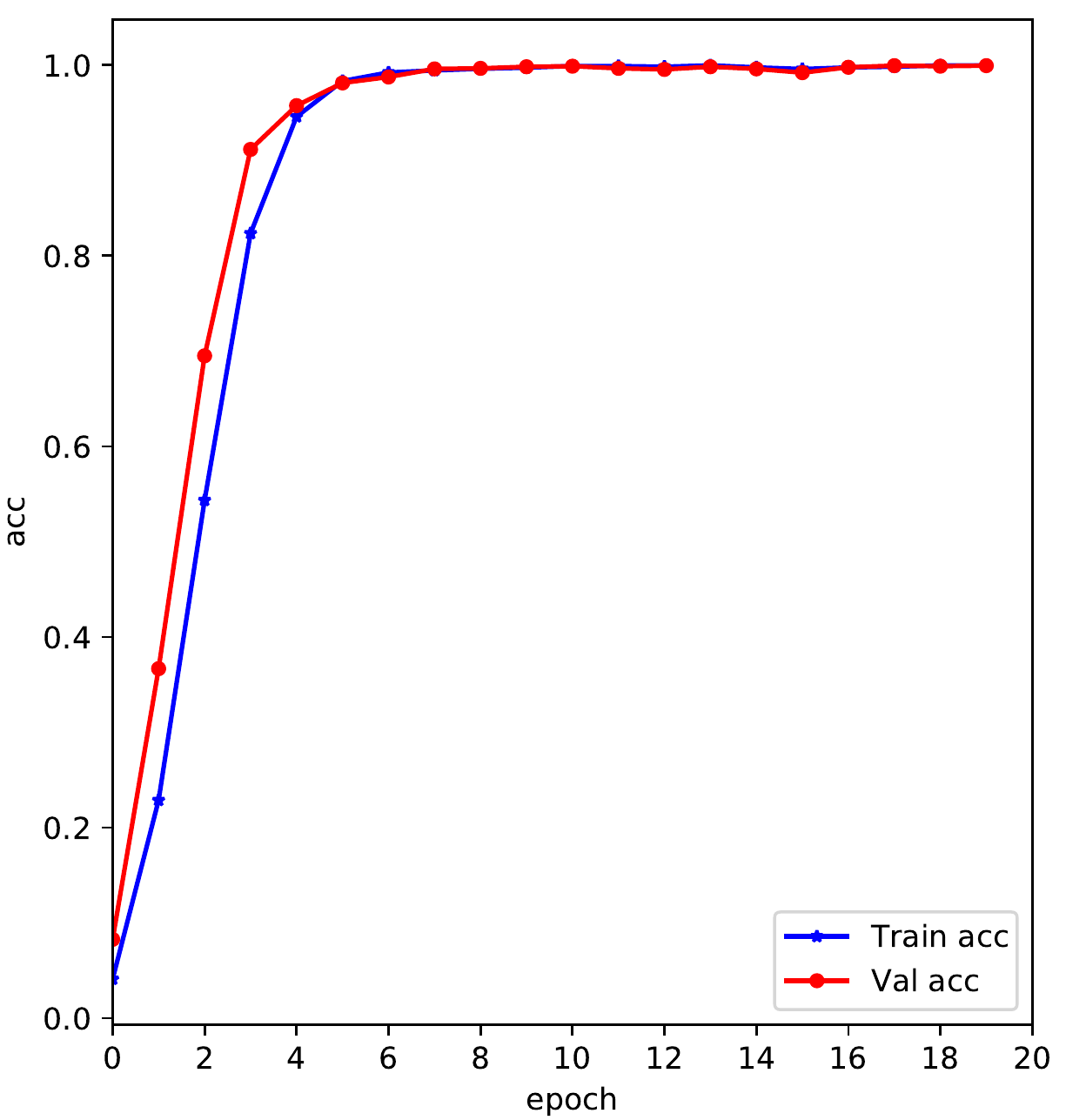} \label{fig-ub}
\end{minipage}
}
\caption{Accuracy curves on two public databases: (a) HKPU, (b) USM.} \label{fig-accuracy}
\end{figure}

In this subsection, we will first show our training process and the results obtained. Fig. \ref{fig-loss} shows our loss curve for training on two publicly available data, and Fig. \ref{fig-accuracy} shows the accuracy curve. The higher number in the number of epochs usually allows the network to be well trained so that the weights of the different layers can be updated accurately. Only 20 epochs are considered for all our experiments, but we can still observe encouraging results Fig. \ref{fig-accuracy} that our validation set accuracy can reach $100\%$, which shows the effectiveness of our method.

In addition, both the proposed network and the basic network were tested on the HKPU database. Table \ref{ta-2} shows the recognition accuracy of the single network (basic network) and our method on HKPU. The basic network uses CNN as the only method to extract the overall features of finger veins. The accuracy of this method is $98.11\%$. In comparison, the proposed network achieved $100\%$ recognition rate. The time in the table is the total time spent for training 20 epochs plus validation, and the difference in time between the two methods is $13$ seconds, indicating the lightweight nature of the proposed model. Overall, the network with the convolutional block attention module in the basic CNN framework improves the recognition performance.
\begin{table}[!h]
\centering
\caption{Comparison results with basic CNN} \label{ta-2}
\begin{tabular}{ccccc}
  \toprule
  Architecture & \makecell{Accuracy} & Time(Training + verification) \\
  \midrule
  Basic CNN & $98.11\%$ & $670$s\\
  Our approach & $100\%$ & $683$s\\
  \bottomrule
\end{tabular}
\end{table}
\subsection{Comparison with the Existing Methods}
Finger vein recognition technology has been developed for more than a decade and several advanced algorithms have been proposed. In this experiment, we have divided the existing methods into traditional methods and deep learning methods and compared them with our proposed method.

Table. \ref{ta-3} shows the results of the comparison with some classical traditional finger vein recognition algorithms. The results show that our scheme significantly outperforms some algorithms, e.g., literature \cite{A-Kumar-2012},\cite{J-Wang-2013},\cite {Y-Xin-2012},\cite{N-Miura-2007},\cite{N-Miura-2004},\cite{L-Yang-2014}, while there is almost no difference in recognition accuracy compared to scheme \cite{L-Yang-2018},\cite{H-C-2010}, but our method does not require additional processing of finger vein images, which greatly reduces the time cost.
\begin{table}[!h]
\centering
\caption{Comparison results with traditional methods}\label{ta-3}
\begin{tabular}{ccc}
\hline
Database  & Method & Accuracy \\ \hline

\multirow{6}{*}{HKPU} & LBP \cite{L-Yang-2014} & $83\%$ \\
                  & Gabor \cite{A-Kumar-2012} & $90.08\%$ \\
                  & MC \cite{N-Miura-2007} & $85.24\%$ \\
                  & RLT \cite{N-Miura-2004} & $78.28\%$ \\
                  & \makecell{Weighted vein indexing \cite{L-Yang-2018}} & $99.68\%$ \\
                  & This Paper & $100\%$ \\ \hline
\multirow{6}{*}{USM} & WLBP \cite{H-C-2010}& $93.8\%$ \\
                  & PCA \cite{J-Wang-2013} & $94.7\%$ \\
                  & SR \cite{Y-Xin-2012} & $95.1\%$ \\
                  & \makecell{Weighted vein indexing \cite{L-Yang-2018}} & $99.93\%$ \\
                  & CPBFL-BCL \cite{H-Y-2020} & $99.98\%$ \\
                  & This Paper & $100\%$ \\ \hline
\end{tabular}
\end{table}

\begin{table}[!h]
\centering
\caption{Comparison results with neural network methods}\label{ta-crnnm}
\begin{tabular}{ccccccc}
\hline
\multirow{2}{*}{} & \multirow{2}{*}{Method} & \multicolumn{2}{c}{Accuracy} & \multirow{2}{*}{\makecell{Number of\\convolution}} & \multirow{2}{*}{\makecell{Number of\\ filters}} & \multirow{2}{*}{Kernel size} \\ \cline{3-4}
                  &  & \multicolumn{1}{c}{HKPU} & USM &  &   &  \\ \hline
\multirow{8}{*}{CNN} & \emph{Hong et al.} \cite{H-G-Hong-2017}   & \multicolumn{1}{c}{$-$} & $95.83\%$ & $13$               & \makecell{$64$, $128$,\\ $256$, $512$} &   $3\times3$   \\ \cline{2-7}
                  & \makecell{CNN with \\original images \cite{R-Das-2019}} & \multicolumn{1}{c}{$95.3\%$} & $97.53\%$ &  \multirow{3}{*}{$5$} & \multirow{3}{*}{\makecell{$153$, $512$,\\ $768$, $1024$}}& \multirow{3}{*}{\makecell{$5\times5$, $4\times15$,\\ $1\times1$}}  \\ \cline{2-4}
                  & \makecell{CNN with CLAHE\\ enhanced images \cite{R-Das-2019}} & \multicolumn{1}{c}{$94.37\%$} & $97.05\%$ &    &     &    \\ \cline{2-7}
                  & Improved CNN \cite{J-Y-2020} & \multicolumn{1}{c}{$-$} & $98\%$ &   $3$   & $8$, $16$, $32$ & $5\times5$, $3\times3$, $3\times3$ \\ \cline{2-7}
                  & \makecell{Two-stream \\ CNN \cite{Y-Fang-2018}} & \multicolumn{1}{c}{$-$} & $95.45\%$ & $3$, $3$ & \makecell{Net1: 20, 50, 200\\Net2: 20, 50, 200} & \makecell{Net1: $9\times9$, $10\times26$\\Net2: $7\times7$, $4\times7$} \\ \cline{2-7}
                  &   Basic CNN       & \multicolumn{1}{c}{$98.11\%$} & $-$ & $2$  & $16$, $32$     &  $5\times5$  \\ \hline
\multirow{5}{*}{Other NN} & \makecell{Deep Belief\\ Network CGN} & \multicolumn{1}{c}{$-$} & $97.4\%$ & \multirow{4}{*}{$-$} & \multirow{4}{*}{$-$} & \multirow{4}{*}{$-$}  \\ \cline{2-4}
                  & AlexNet   & $-$ & $92.28\%$ &    &     &    \\ \cline{2-4}
                  & ResNet-18 & $-$ & $96.04\%$ &    &     &    \\ \cline{2-4}
                  & CAE+CNN \cite{B-Hou-2020}  & $-$ & $99.49\%$ &    &     &    \\ \cline{2-7}
                  & Our method  & \multicolumn{1}{c}{$100\%$} & $100\%$ & $2$ & $16$, $32$ & $5\times5$  \\ \hline
\end{tabular}
\end{table}

Compared with classical networks, such as AlexNet ResNet-18, Basic CNN, the reason for the high accuracy of our scheme is that the weights are assigned adaptively through the attention mechanism, which highlights the important detail information of the image and the extracted features are more distinguishable. The network layer setup in literature \cite{J-Y-2020} is similar to our model structure, but it needs to extract curvature by Gaussian template and use certain methods to extract feature images as input to CNN, which is time costly and computationally not small. Other literature, such as  \cite{H-G-Hong-2017},\cite{Y-Fang-2018},\cite{R-Das-2019}, shows from the network structures listed in Table \ref{ta-crnnm} that the structures of these schemes are not as simple as those of the proposed scheme. Combining Table \ref{ta-2} and Table \ref{ta-crnnm}, it is obvious that our scheme is less computationally intensive and takes less time to train. From the experimental data, the ideas and methods of this paper are completely correct and effective.

\section{Conclusion}
According to the characteristics of finger veins, this paper adds CBAM to the basic CNN framework for finger vein recognition.The CBAM block further refines the features extracted by CNN to make the features more distinguishable. Compared with the classical network model, our model extracts features that better reflect image information with higher accuracy, which has obvious advantages. In the model, we use a simple network structure with lightweight features, less computation and shorter training time. To make our model more practical, in future work, we will consider larger datasets, i.e., in fusing multiple publicly available datasets into one dataset for learning, which makes the model more robust due to the difference in image quality.

\end{document}